%% file: main.tex
\documentclass[sigconf]{acmart} 
\usepackage{wrapfig}
\usepackage{tikz,lipsum}
\usepackage[most]{tcolorbox}
\usepackage{multirow}
\AtBeginDocument{%
  \providecommand\BibTeX{{%
    \normalfont B\kern-0.5em{\scshape i\kern-0.25em b}\kern-0.8em\TeX}}}

\usepackage{caption}
\usepackage{subcaption}

\newcommand{\cnote}[1]{\textcolor{red}{$\ll$\textsf{#1 --Chen}$\gg$}}

\newcommand{\redd}[1]{{#1}} 

\newcommand{\remove}[1]{}

\begin{document}

\title{The Lean Data Scientist:\\
Recent Advances towards Overcoming the Data Bottleneck}

\author{Chen Shani}
\affiliation{%
  \institution{The Hebrew University of Jerusalem}}
\email{chen.shani@mail.huji.ac.il}

\author{Jonathan Zarecki}
\affiliation{%
  \institution{}}
   \email{jonathan.zarecki@gmail.com}
 
\author{Dafna Shahaf}
\affiliation{%
  \institution{The Hebrew University of Jerusalem}}
\email{dshahaf@cs.huji.ac.il}

\renewcommand{\shortauthors}{Shani, et al.}
\newcommand{\xhdr}[1]{\vspace{1mm}\noindent{{\bf #1.}}} 

\begin{abstract}


Machine learning (ML) is  revolutionizing the world, affecting almost every field of science and industry. Recent algorithms (in particular, deep networks) are increasingly \emph{data-hungry}, requiring large datasets for training. Thus, the dominant paradigm in ML today involves constructing large, task-specific datasets.

However, obtaining quality datasets of such magnitude proves to be a difficult challenge. A variety of methods have been proposed to address this \emph{data bottleneck} problem, but they are scattered across different areas, and it is hard for a practitioner to keep up with the latest developments. In this work, we propose a \emph{taxonomy} of these methods. Our goal is twofold: (1) We wish to \emph{raise the community’s awareness} of the methods that already exist and encourage more efficient use of resources, and (2) we hope that such a taxonomy will contribute to our understanding of the problem, inspiring \emph{novel ideas and strategies} to replace current annotation-heavy approaches.  

\end{abstract}

\maketitle

\section{Introduction}
\input{sections/intro}

\label{sec:intro}

\section{Taxonomy}
\input{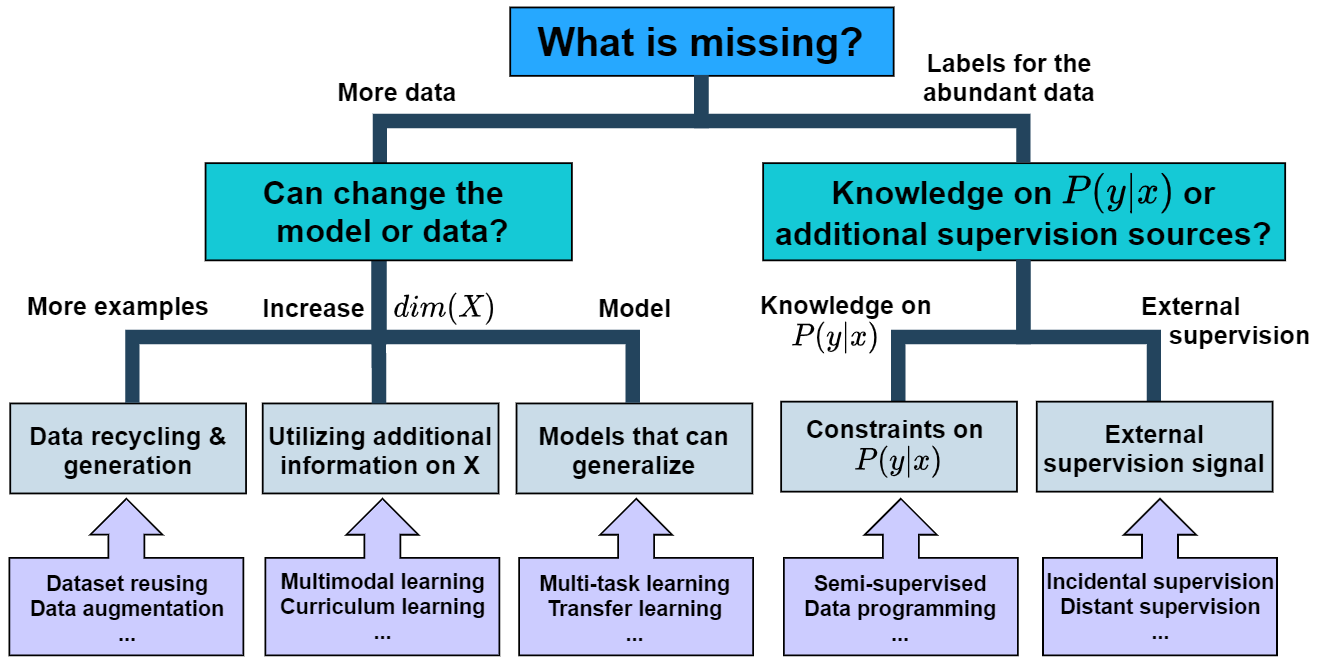}
\label{sec:taxonomy}

\section{Obstacle: Missing Data}
\input{sections/missing_data}

\label{sec:missing_data}

\section{Obstacle: Missing Labels}
\input{sections/missing_labels}
\label{sec:missing_labels}


\section{Discussion \& Conclusions}
\input{sections/conclutions}
\label{sec:conclutions}

\section{Acknowledgments}
We thank the anonymous reviewers, as well as Daphna Weinshall and Guy Hacohen for their insightful comments. 

This work was supported by a grant from Israel Ministry of Science and Technology and by the European Research Council (ERC) under European Union's Horizon 2020 research and innovation program (grant no. 852686, SIAM).

\newpage

\bibliographystyle{ACM-Reference-Format}
\bibliography{bibliography}


\end{document}

%% file: sections/intro.tex
\begin{quotation}
``I think AI is akin to building a rocket ship. You need a huge engine and a lot of fuel. If you have a large engine and a tiny amount of fuel, you won’t make it to orbit. If you have a tiny engine and a ton of fuel, you can’t even lift off. To build a rocket, you need a huge engine and a lot of fuel.
The analogy to deep learning is that the rocket engine is the deep learning models, and the fuel is the huge amounts of data we can feed to these algorithms.'' 

\textit{-- Andrew Ng}

\end{quotation}

\remove{
\begin{figure}[b]
\includegraphics[width=\linewidth]{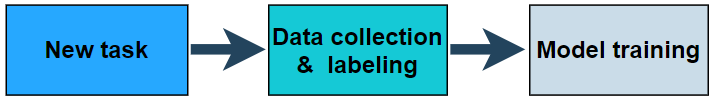}
\addlinespace
\begin{tabular}{p{0.14\textwidth}|p{0.32\textwidth}}
\textbf{Task} & \textbf{Datasets} \\
\bottomrule
Object detection & ImageNet, Visual Genome, BSDS500,  COCO, SUN, Statlog, 
Berkeley 3-D Object, Open Images, Caltech-256, LabelMe \\
\hline
Face \slash \ emotion recognition & RAVDESS, Yale Face Database, UOY 3D-Face, CASIA, FERET, SCFace, BioID Face Database \\
\hline
Action recognition & TV Human Interaction, MHAD, THUMOS, MEXAction2\\
\hline
Handwriting detection & MNIST, Chars74K, UJIPenCharacters, Omniglot, HASYv2 \\
\hline
Sentiment analysis & Amazon reviews, OpinRank, MovieLens, Twitter100k, 
Skytrax, Sentiment140, Lexicoder, OpinRank, Panic! \\
\hline
Dialogues & NPS, Twitter Triple, UseNet, NUS, Reddit,  Ubuntu \\
\hline
Spam detection & Enron, Ling, SMS Spam Collection, Spambase
\end{tabular} 
\caption{Machine learning dominant paradigm with several examples.}
\label{fig:ML_paradigm}
\end{figure}
}

\begin{figure*}[h!]
\includegraphics[width=0.9\linewidth]{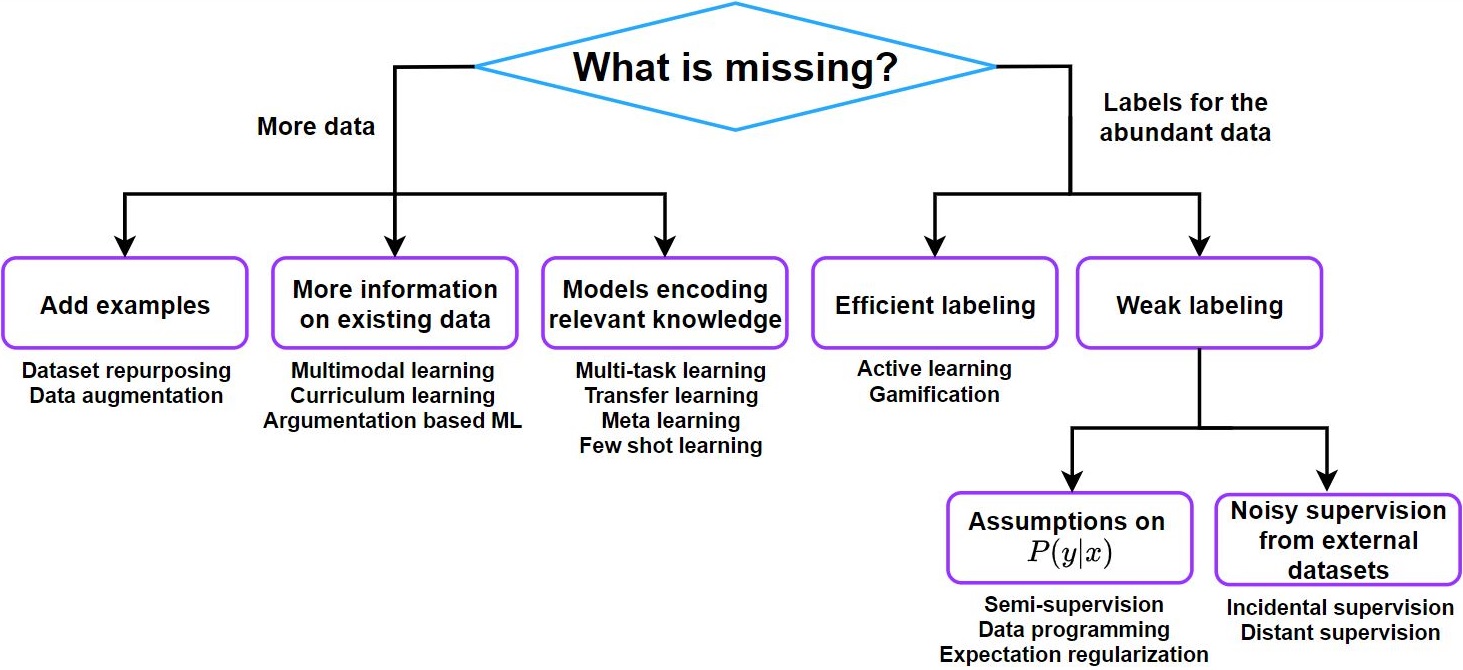}
\caption{\label{fig:taxonomy} Flowchart of the taxonomy for ways to tackle the data bottleneck.}
\end{figure*}

Obtaining data has become the key bottleneck in many machine learning (ML) applications. The rise of deep learning has further exacerbated this issue. Although high-quality ML models are finally making the transition from expensive-to-develop, highly specialized code to something more like a commodity, these models involve millions (or even billions) of parameters and require massive amounts of data to train. Thus, the dominant paradigm in ML today is to create a new (large) dataset whenever facing a novel task (e.g., \citep{rajpurkar2016squad, zellers2018swag, Shorten2019}). In fact, there are now entire conferences dedicated to the creation of new data resources (e.g., LREC, the International Conference on Language Resources and Evaluation \citep{Irec} or resource papers at CIKM \citep{cikm}). 

While this approach resulted in significant advances, it suffers from a major caveat, as collecting large, high-quality datasets is often very demanding in terms of time and human resources \citep{whang2020data}. For several tasks, such as rare disease detection, large datasets are nearly infeasible to construct  \citep{hekler2019we, shu2018small}.

While there has been much effort suggesting workarounds to this \emph{data-bottleneck} problem, they are scattered across many different sub-fields, often unaware of one another. \redd{There exist many method-specific and domain-specific surveys, but broader, big-picture surveys are hard to find. The closest in spirit to our work is \citet{roh2019survey}, which focuses more on the data management point of view and the early stages of the pipeline.}

\redd{In this paper, we aim to bring order to this area. Our main contribution is a simple yet comprehensive taxonomy of ways to tackle the data bottleneck. We survey major research directions and organize them into a taxonomy in a way designed to be useful for practitioners choosing between different approaches. The emphasis here is not on covering methods in depth; rather, we discuss the main ideas behind various methods, the assumptions they make and their underlying concepts. For each topic, we mention several important or interesting works, and refer the interested reader to surveys where possible.}

We wish to first {\bf raise awareness} of the methods that already exist, to encourage more efficient use of data. In addition (and perhaps more importantly), we hope the organization of the taxonomy would \redd{also
reveal gaps in current techniques} and {\bf suggest novel directions of research} that could inspire the creation of new, less data-hungry learning methods. 

%% file: sections/taxonomy.tex
\xhdr{A note on scope} The data-bottleneck problem is widespread across the field of machine learning. It is especially crucial in supervised learning, but applies to unsupervised paradigms as well. In this work we focus on the supervised, unsupervised and semi-supervised settings. Reinforcement learning is generally beyond the scope of this paper, although some of the methods we present are applicable to it. 

\smallskip

We start with a high-level view of our taxonomy, depicted in Figure \ref{fig:taxonomy}. We first make the distinction between cases where \emph{data} ($\mathcal{X}$) is hard to collect, and cases where \emph{labels} ($\mathcal{Y}$) pose the difficulty. 
For example, collecting a dataset of patients with rare diseases is challenging due to the condition's rarity. In contrast, it is relatively easy to collect a large dataset of unlabeled images for an image segmentation task, but annotation is slow and costly. 

If obtaining data is the main obstacle, we identify three major approaches: 
\begin{itemize}
    \item \textbf{Add examples:} Generate more examples from available data (e.g., through data augmentation). 
    \item \textbf{Use additional information on existing data:} Increase the dimensionality of $\mathbf{\mathcal{X}}$ in a manner that can assist the learner (e.g., curriculum learning). 
    \item \textbf{Use models encoding relevant knowledge:} Instead of learning from scratch, take advantage of models trained in a different yet relevant setup (e.g., transfer learning).
\end{itemize} 

If unlabeled data is abundant but labels are difficult to obtain, we identify two main approaches:
\begin{itemize}
    \item \textbf{Acquire labels efficiently:} Label examples that should heavily contribute to the learning process.
    \item \textbf{Weak Labeling:} Using proxy labels, either making assumptions about label distribution (e.g., semi-supervised learning)  or about the labeling process (e.g., data programming), 
    or using external (noisy) supervision signals correlated with the true labels (e.g., incidental supervision).  
\end{itemize}

\redd{We note that these approaches may also be combined. For example, one might add more examples and also increase the dimensionality of the data.}

In the rest of this paper, we follow the taxonomy and elaborate on the different approaches and best practices.

%% file: sections/missing_data.tex
Quite often, data is hard (or impossible) to obtain. 
In the following, we survey some of the main methods from the left branch in Figure \ref{fig:taxonomy}: obtaining more examples efficiently, adding informative dimensions to existing data, or taking advantage of related tasks. 


\subsection{Add Examples}
\label{sec:data_gen}

This category focuses on methods for obtaining more examples. 

\begin{tcolorbox}[colback=blue!2!white,colframe=blue!50!black,title=Dataset repurposing]
Use a preexisting dataset for a new purpose.
\end{tcolorbox}

\textbf{Dataset repurposing} is perhaps the most obvious method to add data, and is mentioned here for the sake of completeness. The idea is to use a preexisting dataset for a different task than it was originally constructed for.  

For example, ImageNet was originally made and used for classification \citep{deng2009imagenet}, but later on was reused for image generation \citep{Oord2016PixelRN}. Similarly, the MS-COCO image captioning dataset \citep{Lin2014MicrosoftCC} was reused for training visually grounded word embeddings \citep{Kottur2015VisualWord2VecL}.

Data repurposing also includes \emph{transformations} on existing datasets. For example, consider \emph{inpainting}, the process of restoring lost parts of an image based on the surrounding information. 
Inpainting is done using various preexisting datasets such as CelebA, Place2 and ImageNet \citep{shin2020pepsi++}, where the same image splits into both $X$ and $Y$ (sometimes in more than one way).

Of course, it is also possible to repurpose a dataset  created with no machine-learning task in mind at all: for example, \citet{bertero2016predicting, bertero2016deep} used a dataset of TV sitcoms for a supervised humor detection task, with recorded laughter serving as labels.



\begin{tcolorbox}[colback=blue!2!white,colframe=blue!50!black,title=Data augmentation]
Perform transformations on $\mathcal{X}$ to enlarge the dataset.
\end{tcolorbox}

\textbf{Data augmentation} is a common approach for generating more data; it artificially inflates the training set by applying modifications \citep{shorten2019survey}. This method's initial goal was to prevent overfitting. 

\begin{figure}[ht]
	\centering
  	\includegraphics[width=\linewidth]{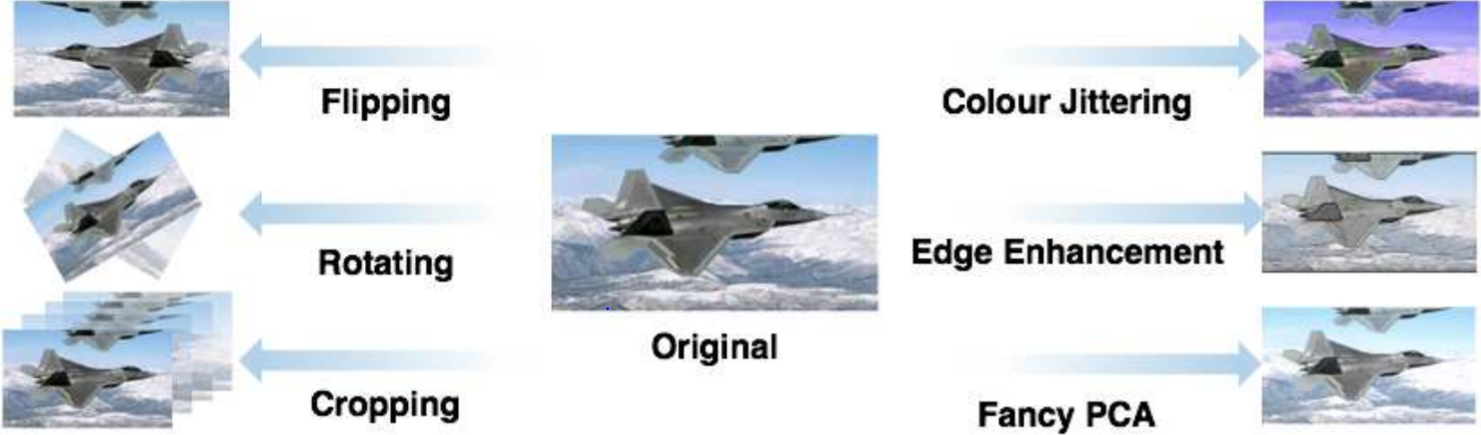}
  	\caption{Examples for common data augmentation manipulations of images as presented by \citet{taylor2017improving}.}
  \label{fig:dataAug}
\end{figure}

Data augmentation often employs \emph{vicinal risk minimization} (VRM) \citep{vapnik2000vicinal, chapelle2001vicinal}. In VRM, human knowledge is needed to define a neighborhood
around each example in the training data, and virtual examples are drawn from this vicinity distribution. It is easiest to demonstrate this idea in the field of computer vision; there, common augmentations are geometric transformations such as flipping, cropping, scaling and rotating (see Figure \ref{fig:dataAug}). The idea is to make the classifier invariant to change in position and orientation. Similarly, photometric transformations amend the color channels to make the classifier invariant to change in lighting and color. 
%


Data augmentation leads to improved generalization, especially with  small datasets \citep{simard1998transformation, taylor2018improving, wei-zou-2019-eda, anaby2020not} or when the dataset is unbalanced (instead of undersampling, which is data-inefficient).

Augmentation methods have seen a recent surge of interest \citep{Shorten2019}. Recent advances include  
methods 
that jointly train a model for generating augmentations \citep{perez2017effectiveness}, and methods that learn which augmentations best fit the data \citep{Cubuk2019AutoAugmentLA, Wei2019EDAED, Ratner2017LearningTC}. 
For example,  AutoAugment \citep{Cubuk2019AutoAugmentLA}  randomly chooses a sub-policy of batch transformation and searches for the one that yields the highest validation accuracy.  

Beyond human-defined transformations, recent methods suggested using pretrained generative-adversarial-networks (GANs) to create new examples \citep{antoniou2017data, Bowles2018GANAA, Choi_2019_ICCV}.
Interestingly, the generated data points do not have to be \emph{interpretable} by humans. For example, Mixup \citep{Zhang2017mixupBE} trains a neural network on convex combinations of pairs of examples and their interpolated labels, treating it as ``noisy'' training data.



\subsection{More Information on Existing Data}
\label{sec:add_x}

Instead of adding new data points, this set of methods focuses on adding dimensions to existing points.


\begin{tcolorbox}[colback=blue!2!white,colframe=blue!50!black,title=Multimodal learning]
Integrate associated information on $\mathcal{X}$ from multiple modalities. 
\end{tcolorbox}

\textbf{Multimodal learning} attempts to enrich the input to the learning algorithm, giving the learner access to more than one modality of $\mathcal{X}$; for example, an image accompanied by its caption. Multimodal learning is intuitive and similar to how infants learn (i.e., children see new objects is often accompanied by additional semantic information). 
The main drawbacks of multimodal learning are obtaining rich input and effectively integrating it into the model.

Although the term ``multimodal learning'' is fairly new, many works combined information from different modalities \citep{lampert2009learning, farhadi2009describing, tian2018audio}. These works, and more recent papers \citep{Schnfeld2018GeneralizedZA, Luketina2019ASO, changpinyo2016synthesized, qiao2016less}, show the promise of this method as an effective way to reduce data requirements and improve generalization.

\begin{figure}[ht]
	\centering
  	\includegraphics[width=\linewidth]{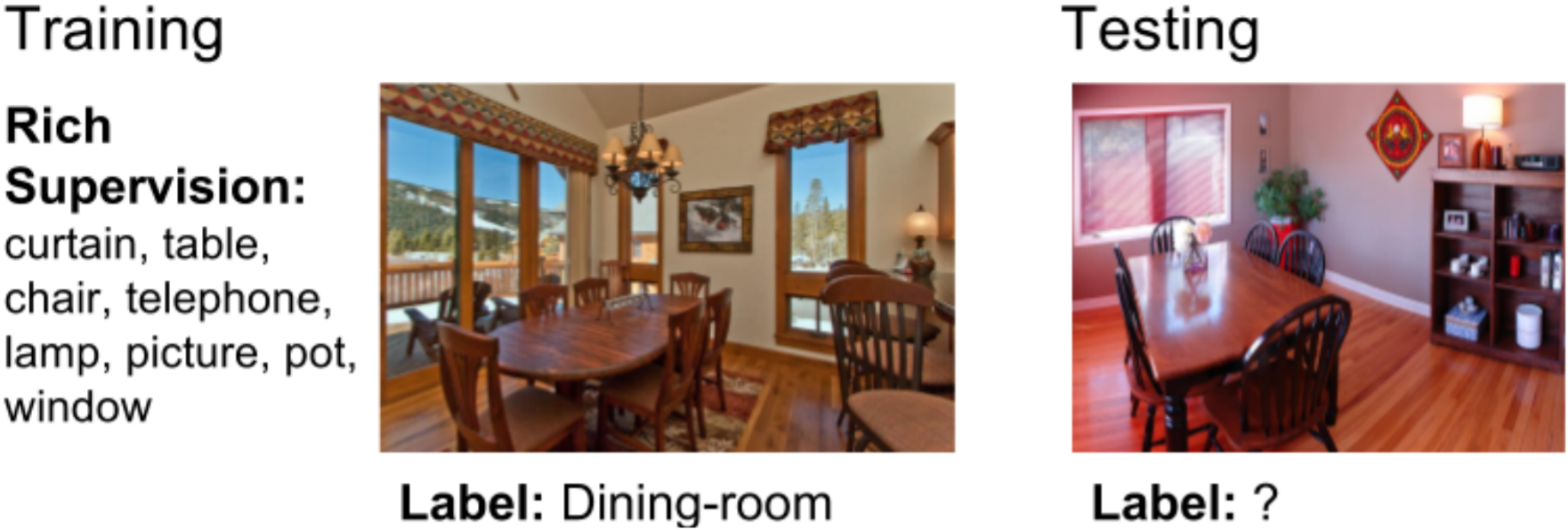}
  	\caption{An illustration of the learning setup used by \citet{visotsky2019few}: labeled examples are accompanied with rich information that provides hints or explains classification. These labels are created during the training phase, where annotators write a list of objects they observe in a visual scene using free text. Irrelevant background objects are often ignored. During the test phase, only the image is provided.}
  \label{fig:richSupervisionChechik}
\end{figure}

Moreover, multimodal learning is also often used when the number of data points is extremely small, and in particular few-, one- and zero-shot learning 
(when only a few target-specific labeled examples exist for the learning process; thus, the learner must understand new concepts using only a handful of examples). 
For example, \citet{visotsky2019few} used multimodal learning for few-shot learning by integrating additional per-sample information -- in this case, a list of objects appearing in the input image (see Figure \ref{fig:richSupervisionChechik}).
\citet{schwartz2019baby} demonstrated that it is possible to outperform previous state of the art results on the popular miniImageNet and CUB few-shot learning benchmarks by combining images with multiple and richer semantics (category labels, attributes and natural language descriptions). 



\begin{tcolorbox}[colback=blue!2!white,colframe=blue!50!black,title=Curriculum learning]
Present examples to the learner according to a predetermined order, usually based on difficulty. 
\end{tcolorbox}

In \textbf{curriculum learning}, the learner is exposed to examples using a predetermined curriculum, where examples are usually sorted in increasing order of difficulty. Meta-data on $\mathcal{X}$ is needed to determine its place in the learning process.  

The motivation behind curriculum learning comes from humans, as teachers tend to start by 
teaching simpler concepts (e.g., learning to ride a bicycle with training wheels first) \cite{avrahami1997teaching}. 
Thus, curriculum learning attempts to augment training examples with a difficulty score, often corresponding to \emph{typicality}. 

\begin{figure}
     \centering
     \begin{subfigure}[b]{0.45\textwidth}
         \centering
         \includegraphics[width=\textwidth]{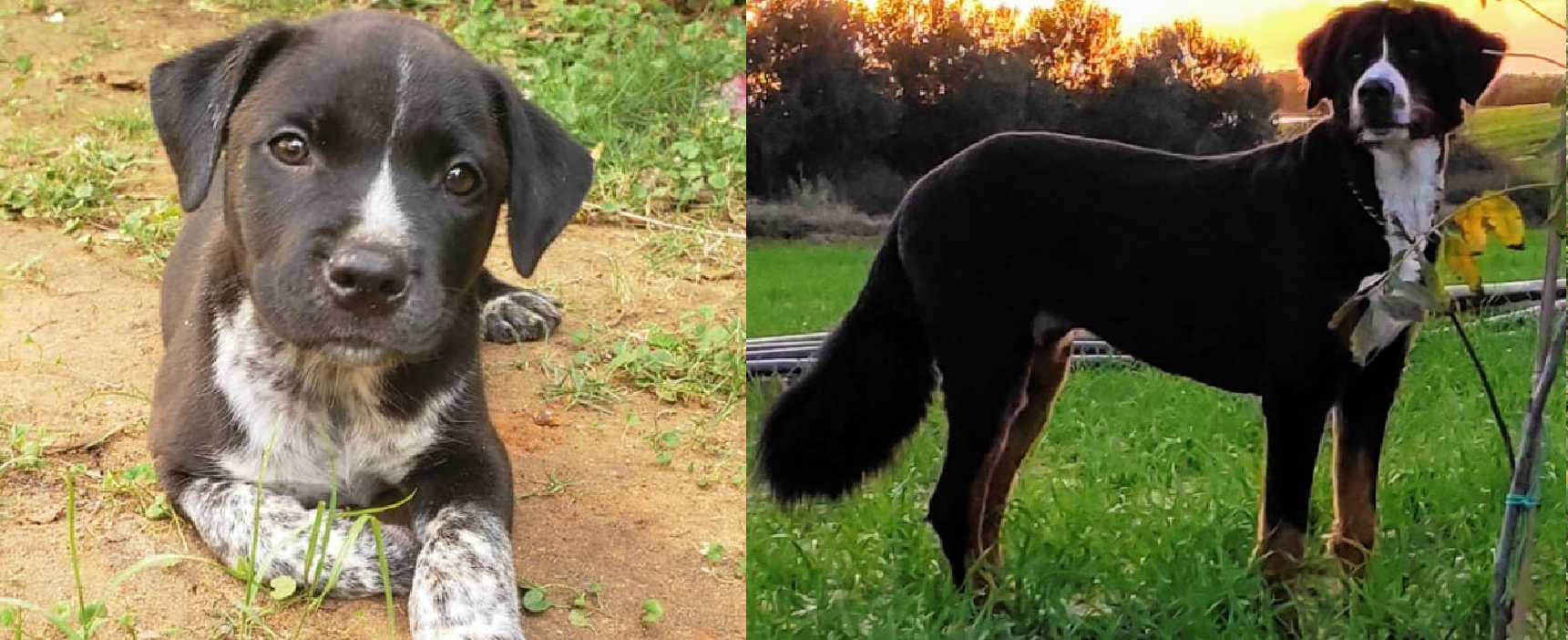}
         \caption{Typical images of dogs, easier to classify.}
         \label{fig:easyCurriculum}
     \end{subfigure}
     \hfill
     \begin{subfigure}[b]{0.45\textwidth}
         \centering
         \includegraphics[width=\textwidth]{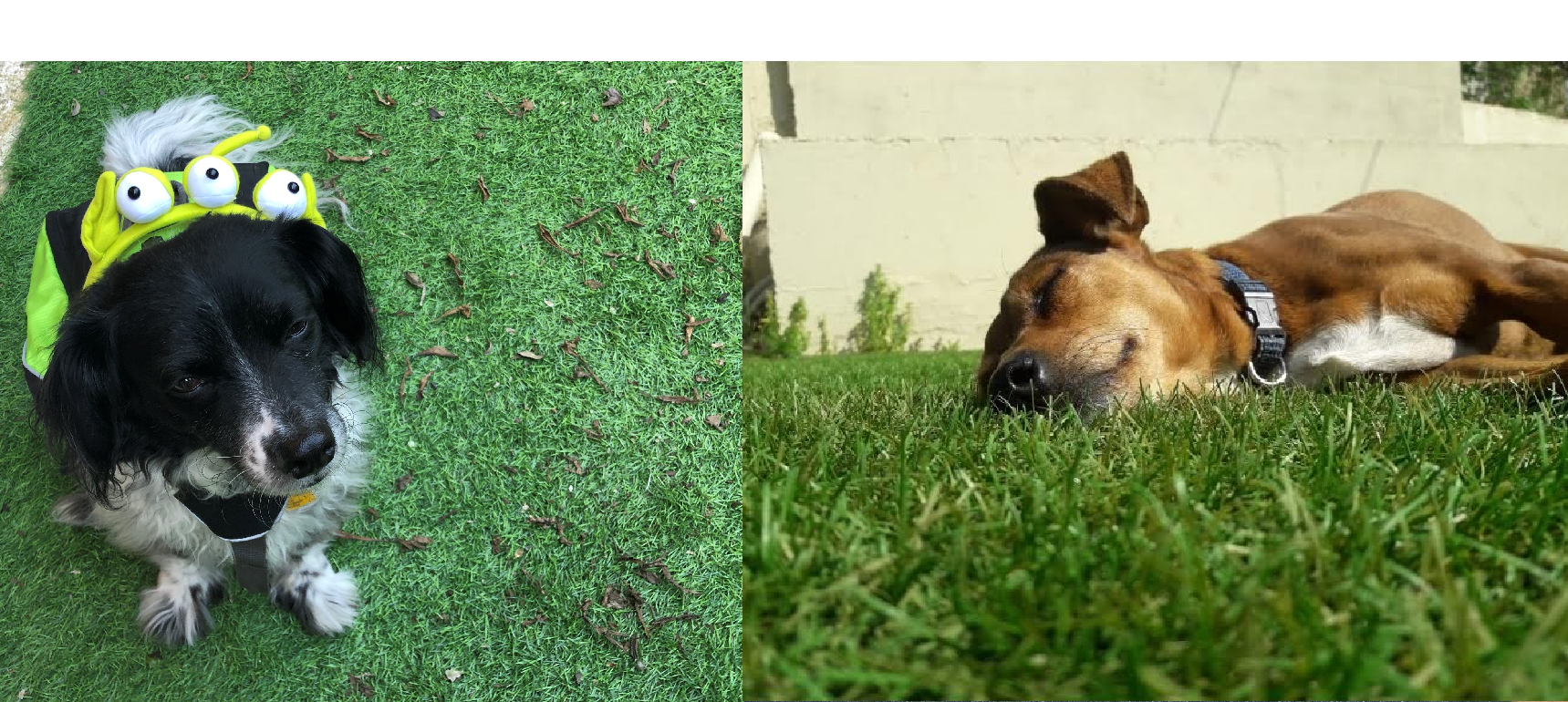}
         \caption{Non-typical images of dogs, harder to classify.}
         \label{fig:hardCurriculum}
     \end{subfigure}
     \caption{Typical versus non typical images of dogs are considered to be easy versus hard, respectively, in a dogs versus cats classification task.}
      \label{fig:examplesCurriculum}    
\end{figure}

Given the difficulty score, the algorithm starts with a set of simple data points and gradually increases the difficulty of training examples throughout the learning process. This progression enables the model to learn the broad concept on a few easy examples and later refine the concept with more difficult ones. 
See Figure \ref{fig:examplesCurriculum} for example; the photos of dogs in the top row are more typical, and should be easier for a classifier to recognize.

Curriculum learning has been shown to improve performance while decreasing the number of examples needed for convergence \citep{bengio2009curriculum, hacohen2019power}. For example, \citet{zaremba2014learning} showed how curriculum improves learning for the task of predicting the output of Python code  without executing it.  

A major caveat of curriculum learning is the inherent need for a difficulty-label estimator \citep{hacohen2019power}. Human labeling of difficulty can be very demanding, perhaps even more than standard annotation. \redd{In practice, the difficulty of each example is often learned by a teacher model, which may have access to related training data \citep{hacohen2019power, matiisen2019teacher, hacohen2020let, huang2020curricularface, forestier2017intrinsically, graves2017automated}.}

\redd{A related concept is \emph{self-paced learning} (SPL) \citep{kumar2010self,jiang2015self}. Intuitively, the curriculum in SPL is determined by the student’s abilities, 
rather than being fixed by the teacher. Instead of heuristically designing a difficulty measure, SPL introduces a regularizor into the learning objective, with the goal of optimizing a curriculum for the model itself. This makes SPL broadly applicable.}



\begin{tcolorbox}[colback=blue!2!white,colframe=blue!50!black,title=Argumentation-based machine learning]
Use experts' local knowledge to restrict the search space. 
\end{tcolorbox}

\textbf{Argumentation-based machine learning} (ABML) is a method to constrain the search space using \emph{experts' local knowledge} \citep{clark1989cn2, movzina2007argument, rahwan2009argumentation, modgil2013added}. 
In a nutshell, in ABML the learner attempts to find if-then rules to explain argumented examples in a rule induction process. 
The learner starts by finding a rule, adding it to a set of rules and removing all training data points that are covered by that rule. This process is repeated until all examples are removed. 
ABML's main advantage is the use of expert knowledge to justify \emph{specific} examples, which is often easier than explaining global phenomena.

For example, \citet{movzina2007argument} used ABML for medical records of deceased patients, where they used a physician’s reasoning for the cause of death to limit the search space. 

ABML is perhaps less popular than the other methods in this section. Nevertheless, if  expert local knowledge is available, ABML is a powerful way to integrate partial prior knowledge. Moreover, the induced hypothesis should make more sense to an expert, as it must be consistent with the input arguments. 


\subsection{Models Encoding Relevant Knowledge} 
\label{sec:improve_model}

In this part, we go beyond the classical pipeline of training a model for a task; we present models that can take advantage of other, related tasks.



\begin{tcolorbox}[colback=blue!2!white,colframe=blue!50!black,title=Multi-task learning]
Co-learn multiple tasks simultaneously to enhance cross task similarities for better generalization.   
\end{tcolorbox}


\textbf{Multi-task learning} (MTL) is a prominent area of research where one attempts to train on multiple different (yet related) tasks simultaneously. These multiple tasks are solved concurrently, exploiting commonalities and differences across them. 

It has been shown that challenging the learner to solve multiple problems at the same time results in better generalization and better performance on each individual task \citep{Ruder2017AnOO}. Indeed, MTL is successfully used in both vision and NLP \citep{kendall2018multi, Liu2019MultiTaskDN}.
The key factors for this success in the absence of a large dataset are: (1) It is an implicit data augmentation method, based on cross-task commonalities; (2) It enables unraveling cross tasks and feature correlations; (3) Encouraging a classifier to also perform well on a slightly different task is a better regularization than uninformed regularizers (e.g., enforce weights to be small, which is the typical $L2$-regularization).

As an example, consider the case of spam-filtering. Quite often, data from an individual user is insufficient for training a model.
Intuitively, different people have different distributions of features that distinguish spam from legitimate email. For example, emails in Russian are probably spam for English speakers, but not for Russian speakers. However, inter-user commonalities can be utilized to solve this problem (e.g., text related to money transfer is probably spam). To build upon these similarities, \citet{attenberg2009collaborative} created an MTL-based spam-filter, treating each individual user as one distinct but related classification task and training a model across the different users.

\begin{figure}[ht]
	\centering
  	\includegraphics[width=\linewidth]{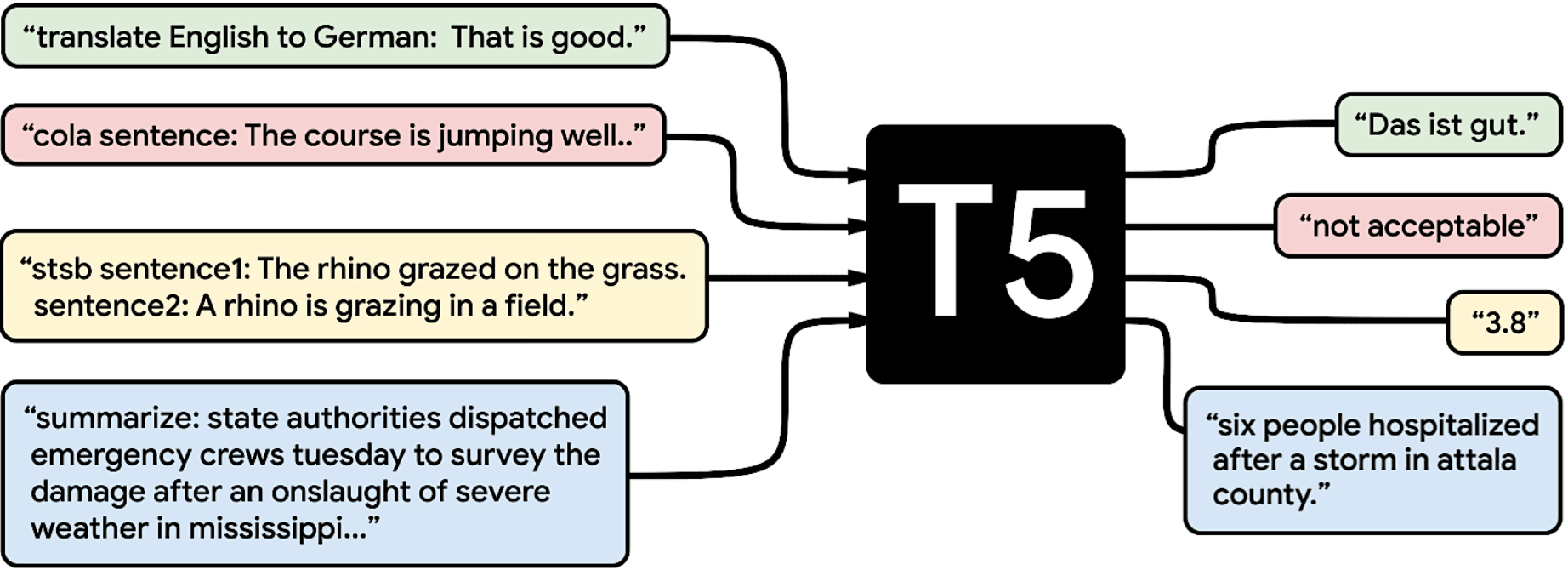}
  	\caption{Multi-task paradigm as presented by \citet{raffel2019exploring}: the objective is to train a model to perform several tasks that are closely related. The input contains the current task, which allows the use of same model, loss function and hyperparameters across various tasks.}
  \label{fig:multiTask}
\end{figure}

A more recent example of MTL learning is the T5 model (see Figure \ref{fig:multiTask}) \citep{raffel2019exploring}. This model achieves state-of-the-art results on many NLP benchmarks while being flexible enough to be fine-tuned to a variety of downstream tasks. T5 receives as input the task at hand and thus allows the use of the same model, loss function, and hyperparameters for any NLP task.

MTL implementations can be divided into two main categories -- hard versus soft parameter sharing of the hidden layers, where hard parameter sharing is more commonly used. In the hard type, the hidden layers are shared between all tasks while keeping several task-specific output layers. \citet{baxter1997bayesian} showed that hard parameter sharing reduces the risk of overfitting to order N (the number of tasks), which is smaller than the risk of overfitting the task-specific parameters (the output layers). 
In soft parameter sharing, each task has its own model and parameters. The distance between model's parameters is then regularized to encourage them to be similar (enhance cross tasks’ similarity), as done by \citet{duong2015low}. 


\remove{
MTL can also occur when training on diverse datasets where a single example might not have labels on all the sub-tasks, but a single model is used to jointly solve them all \citep{Kokkinos2016UberNetTA, Liu2019MultiTaskDN}. 
Such models learn a shared representation for all tasks, enabling them to perform well even on tasks with relatively small datasets. 
}

\remove{
\begin{figure}[b]
     \centering
     \begin{subfigure}[b]{0.42\textwidth}
         \centering
         \includegraphics[width=\textwidth]{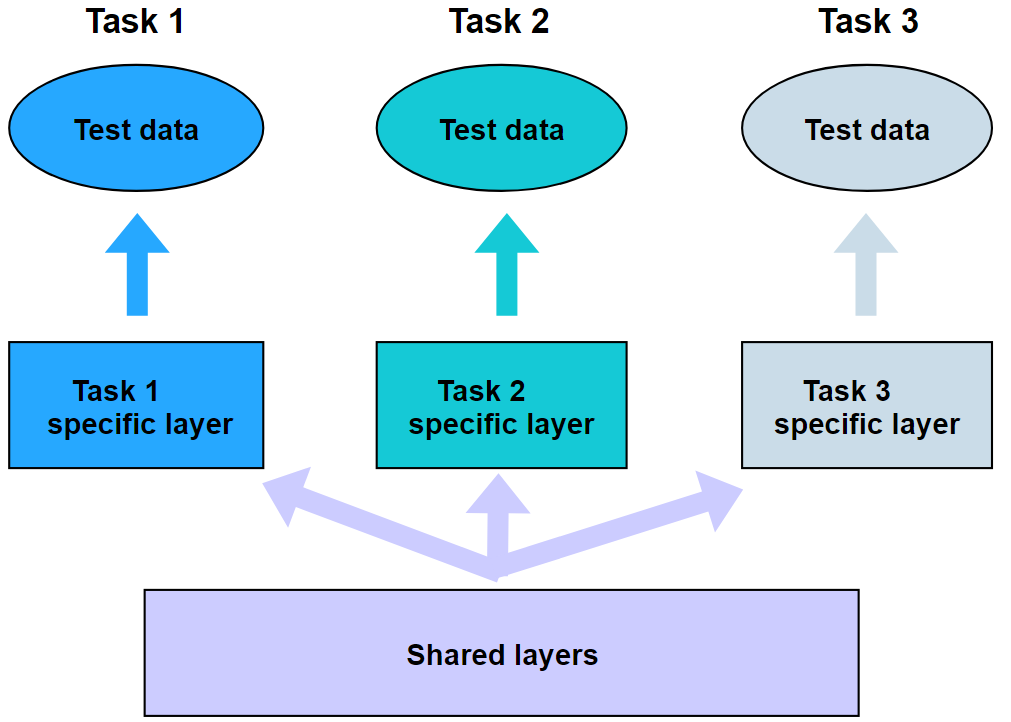}
         \caption{Hard parameter sharing in the MTL paradigm. The hidden layers are shared across the different tasks, mathematically equivalent to strong $L_0$ regularization.}
         \label{fig:MTLHard}
     \end{subfigure}
     \hfill
     \addlinespace
     \begin{subfigure}[b]{0.42\textwidth}
         \centering
         \includegraphics[width=\textwidth]{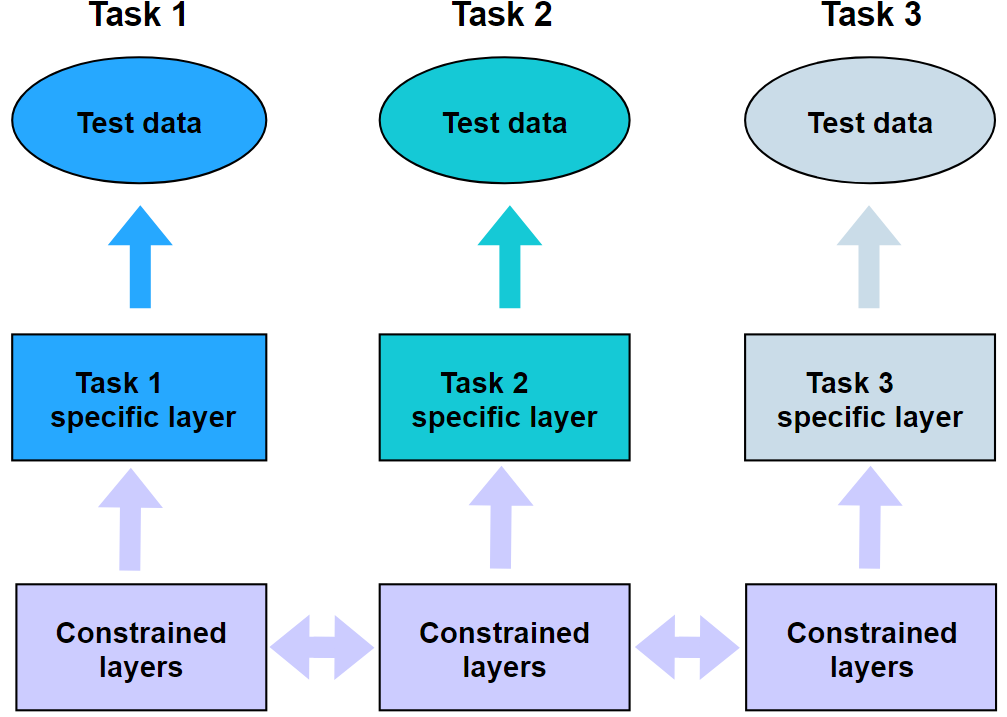}
         \caption{Soft parameter sharing in the MTL paradigm. Each task has separate parameters and hidden layers. Usually the loss at the outer layer is regularized by the current distance between the models.}
         \label{fig:MTLSoft}
     \end{subfigure}
     \caption{Hard versus soft parameter sharing in MTL.}
      \label{fig:MTLHardSoft}    
\end{figure}
}


\begin{tcolorbox}[colback=blue!2!white,colframe=blue!50!black,title=Transfer learning]
Transfer knowledge gained while solving one problem to a different yet related problem.
\end{tcolorbox}

\textbf{Transfer learning} is a widely used, highly effective way to integrate prior knowledge, similar to humans, who never approach a new problem tabula rasa, but rather with rich experience of somewhat similar problems and their solutions \redd{\citep{pan2009survey,torrey2010transfer}}.

The idea is to use preexisting models trained on related tasks. These pretrained models are usually used as an initialization for fine-tuning using a small dataset for the task in hand.
Thus, significantly less task-specific examples are needed for convergence. 

Another beneficial side effect is the use of the model's initial wide domain knowledge, compared to initialization with random weights. In other words, the model starts the fine-tuning phase with some  relevant world knowledge.

\remove{
\begin{figure}
     \centering
     \begin{subfigure}[b]{0.42\textwidth}
         \centering
         \includegraphics[width=\textwidth]{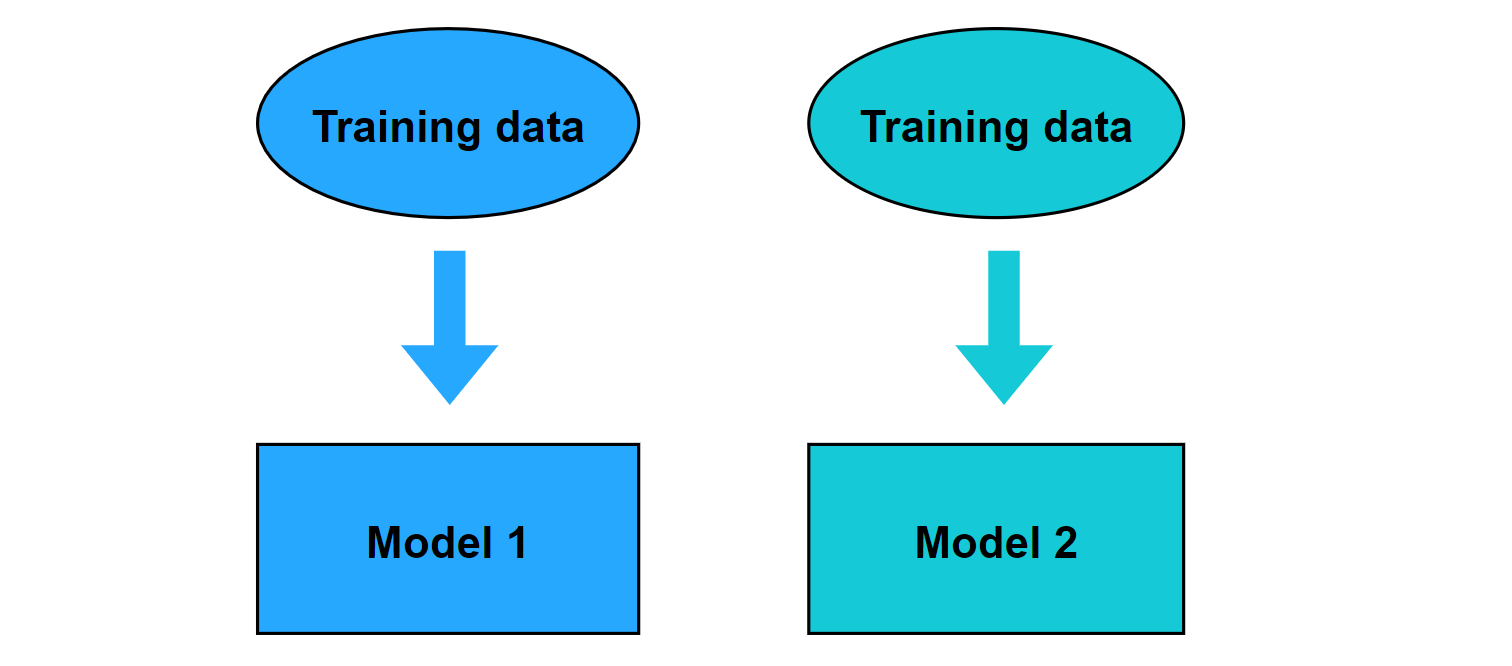}
         \caption{Conventional learning paradigm of models that are limited to the specific task they were created for.}
         \label{fig:ConTransfer}
     \end{subfigure}
     \hfill
     \addlinespace
     \begin{subfigure}[b]{0.42\textwidth}
         \centering
         \includegraphics[width=\textwidth]{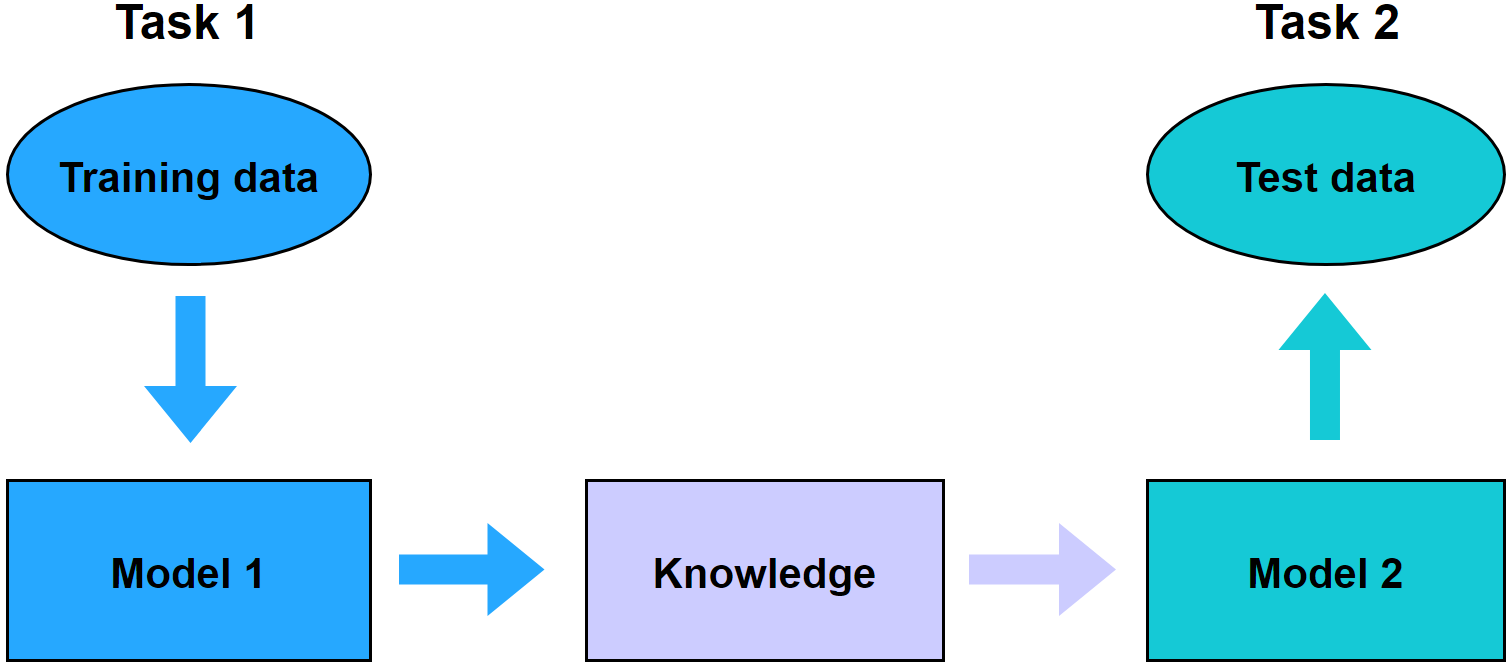}
         \caption{Transfer learning paradigm where a model that gained general knowledge using related task(s) is used for a novel problem.}
         \label{fig:TransferPard}
     \end{subfigure}
     \caption{Transfer versus conventional learning paradigm: the source model is trained on a large related dataset. This model is then used for the low data-budget task.}
      \label{fig:transfer}    
\end{figure}
}

For example, models trained on ImageNet \citep{deng2009imagenet}  have been transferred to medical imaging tasks, including inspecting chest x-rays \citep{wang2017chestx,rajpurkar2017chexnet} and retinal fundus images \citep{gulshan2016development,de2018clinically}. 
The idea is that a network trained on a large and diverse dataset of images captures universal visual features such as curves and edges in its early layers (similar to the primary visual cortex of humans and many other mammals \citep{somers1999functional, vidyasagar2015origins}, a Nobel prize winning discovery\footnote{\url{https://www.nobelprize.org/uploads/2018/06/hubel-lecture.pdf}}). Despite the difference between the images in ImageNet and those in the downstream tasks, these features are relevant for many vision tasks. Therefore, this approach significantly decreases the size of labeled task-specific data needed.

In NLP, the commonly used pretrained model BERT achieves state-of-the-art results in various tasks \citep{devlin-etal-2019-bert}. 
Pretraining such models is often done in a \emph{self-supervised} manner, where different parts of the input are masked, and the learner's goal is to predict the masked parts. For example, given a sentence, it is possible to iterate over it, masking a different word each time, to create various examples.

Fine-tuning in deep networks is usually done either by adding an untrained last layer and training the new model on the small task-specific dataset or by taking the output embeddings of the next to last layer. Another possible fine-tuning technique is to train the whole network with a relatively small learning rate; that is, perform small changes on the already-decent weights (as a heuristic, about ten times smaller than the learning rate used for pretraining). Fine-tuning can also be done by freezing the weights of the first few layers of the pretrained model. The motivation behind this technique is that the first layers capture universal features that would probably also be relevant to the new task. Thus, freezing them during fine-tuning should keep the captured information that is relevant for both the original and the new tasks.

To conclude, transfer learning is a powerful tool for both reducing the amount of task-specific data needed and improving models' performance. 


\begin{tcolorbox}[colback=blue!2!white,colframe=blue!50!black,title=Meta learning]
Improve the learning algorithm by generalizing based on experience from multiple learning episodes.
\end{tcolorbox}

\textbf{Meta-learning} (also known as ``learning to learn'') is a recent subfield of machine learning \citep{finn2017model, Sun2018MetaTransferLF}, focusing on designing models that can learn new tasks or adapt to new environments rapidly, with only a few training examples. It is based on creating a meta-learner that has wide prior knowledge regarding the relevant topic(s). Meta learning is also inspired by human learning. For example, people who know how to ride a bicycle are more likely to quickly learn to ride a motorcycle. 



\redd{Note that while meta learning can often be meaningfully combined with MTL systems, their objectives are different. While MTL aims to solve all training tasks, meta learning aims to use the training tasks for solving \emph{new} tasks with small data. Thus, meta learning is about creating models with \textit{prior experience} that can quickly adapt to new tasks. Specifically, the meta-learner gradually learns meta-knowledge across tasks, which can be generalized to a new task using little task-specific information.}

There are three common approaches to meta-learning: metric-based (similar to nearest-neighbor algorithms), optimization-based (meta-gradients optimizing) and model-based (no assumptions about data distribution). 


As an example of a metric-based approach, \citet{vinyals2016matching} proposed a framework that explicitly learns from a given support set to minimise a loss over a batch. The end result is a model that learns to map a small labeled support set and an unlabelled example to its label, obviating the need for fine-tuning to adapt to new class types. They then showed the superiority of this method in both vision and NLP tasks. 

\remove{
As an example of metric-based approaches, \citet{koch2015siamese} proposed a method to use a Siamese network for one-shot learning. 
Siamese network is a neural network architecture where instead of learning to classify its inputs, the network learns to differentiate between two inputs. \citet{koch2015siamese} used Siamese network for one-shot learning image classification a follows: First, the Siamese network is trained on a verification task, where its output is the probability of two images belonging to the same class. During testing, the network processes all the image pairs between a test image and every image in the support set. The final prediction is the class of the support image with the highest probability.
}

A well-known work in the optimization-based line of research is model-agnostic meta-learning (MAML), which is a fairly general optimization algorithm, compatible with any gradient descent-based model \citep{finn2017model}. It uses a meta-loss specifically designed to induce quick changes when fine-tuned on new tasks, and is based on $N$-gradients (where $N$ is the total number of tasks).

In the model-based line of research, \citet{munkhdalai2017meta} presented MetaNet, a meta-learning model designed specifically for rapid generalization across tasks. The rapid generalization of MetaNet relies on ``fast weights'', which are parameters of the network with a smaller timescale for changes than the regular gradient-based weight changes. This Hebbian short-term plasticity maintains a dynamically changing short-term memory of the recent history of the units' activities in the network, as opposed to the standard slow recurrent connectivity. This model outperforms various other recurrent models across several tasks.

%% file: sections/missing_labels.tex
We now turn our attention to the second major branch in Figure \ref{fig:taxonomy}, where unlabeled data is abundant, but there are few labels (or no labels at all). This setting is common in practice because unlabeled data is often much easier to obtain than labeled data. 
In this section we cover two main approaches.  The first deals with ways to acquire labels efficiently, and the other uses weak labels.

\subsection{Acquiring labels efficiently}
\label{sec:efficient}

\begin{tcolorbox}[colback=blue!2!white,colframe=blue!50!black,title=Active learning]
Generate  examples which are close to the decision boundary. These examples should contribute to the learning process more than random examples.
\end{tcolorbox}

When more labels are needed but annotation is costly, an immediate question would be how to acquire new labeled data \emph{efficiently}. The prime example of this is
active learning, in which the learner can iteratively query an oracle (information source) to label new data points \citep{settles2009active,ren2021survey}. These queries can include unlabeled examples either from the dataset or new ex-nihilo data points, often ones that are close to the decision boundary. The rationale is that not all examples contribute equally to the learning process: diverse examples that are difficult for the learner to classify might be especially useful and could decrease the number of data points needed for learning \citep{Winston1976ThePO}. 




There are many methods to determine which data points from the training set should be queried next. Common objectives include picking examples which will change the current model the most, examples which the current model is least certain about, or diverse examples that resemble the data distribution \citep{settles2009active, beluch2018power, ash2019deep, sener2017active, hacohen2022active}. \remove{In recent years there is a growing interest in integrating semi-supervised methods, showing inconsistent results \citep{}.} For example, \citet{hacohen2022active} recently showed that in the presence of little data it is most beneficial to present the model with typical examples (compared to scenarios with more data, in which it is best to use examples that are close to the decision boundary).

When generating \emph{new} examples (rather than selecting unlabeled ones from the training set), it is important to remember that humans will be the ones labeling them. 
\redd{We wish to point out that while data augmentation modifies the input but \emph{keeps} its label (see Section \ref{sec:missing_data}), active learning generates examples \textit{without labels}. 
Thus, the generation algorithm should keep the new points interpretable, i.e., ensure they have a clear label \citep{nela2006alnearmiss}. 
For example, \citet{zarecki2021textual} automatically transformed sentences' sentiment by replacing key words that bring them closer to the classification boundary (while keeping their syntax). }

Recent approaches use GANs to generate new examples, either from scratch (and label them) \citep{Zhu2017GenerativeAA}, or by modifying an existing example (while attempting to preserve the label) \citep{Tran2019BayesianGA}.
Both scenarios update the learner and the GAN model simultaneously after labeling a new example. 

\begin{figure}[h]
    \centering
    \subfloat[MNIST]{{\includegraphics[width=0.15\textwidth]
    	{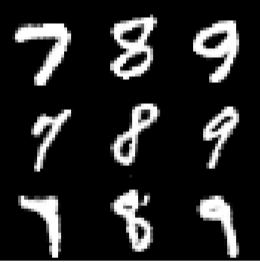} }}
    \quad
    \subfloat[CIFAR10]{{\includegraphics[width=0.15\textwidth]
    	{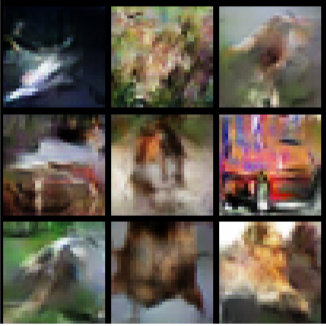} }}
    \caption{Images generated by the GAN transformation approach for ``near-miss'' examples \cite{Tran2019BayesianGA}. Images generated based on the MNIST dataset are interpretable to humans, while this is not the case for the CIFAR10 examples. }
    \label{fig:example_gan_transformations} 
\end{figure}

Importantly, the GAN approaches are more expressive than transformation-based approaches, but the result is often less interpretable. Figure \ref{fig:example_gan_transformations} shows an example of modified images from \citet{Tran2019BayesianGA}. Note that while the MNIST examples (handwritten digits) have relatively clear labels, the CIFAR10 examples (tiny images in ten classes such as airplane, dog, and ship) are not as easy to label. 


\smallskip
\noindent {\bf A note on gamification.} 
Active learning is the dominant paradigm for reducing the \emph{number} of annotations needed. However, a different approach to label efficiently is to reduce the \emph{cost} of annotations. A notable example is \emph{gamification} -- applying gaming mechanics to non-gaming environments, to make tasks more enjoyable and give annotators a non-monetary incentive to provide labels. \redd{The challenge in gamification is often to design the game to create the right incentive. This is far from trivial, and requires 
knowledge of game design, motivational psychology, and an understanding of the target group  \cite{morschheuser2019gamification}. Ignorance of the complexity involved in gamification often results in modest outcomes \cite{landers2019gamification}.}

The seminal work of \citet{von2004labeling} demonstrated a two-player game for image labeling, where the players gain points for describing an image using the exact same term. The researchers famously estimated that if users were to play the game at the same rate as other popular online games, most images on the web could be labeled (for free) within only a few months. Another example is the unfun.me corpus used in humor research. This corpus was constructed via an online game where players change satirical headlines into serious ones with minimal edits \citep{west2019reverse}.


\subsection{Weak labeling}

If we cannot obtain labels efficiently, we could choose to obtain noisy labels as a proxy. \redd{In vision, this is sometimes referred to at ``Automatic Image Annotation'' \cite{cheng2018survey}.} We cover two main types of noisy labels below.

\subsubsection{Assumptions on $P(\mathcal{Y}=y | \mathcal{X}=x)$}
\label{sec:constraints}\ \\ \

\begin{tcolorbox}[colback=blue!2!white,colframe=blue!50!black,title=Semi-supervised learning]
Harness information regarding $P(\mathcal{X}=x)$ to reduce labeling requirements by integrating labeled and non-labeled examples in the learning process. 
\end{tcolorbox}

\textbf{Semi-supervised learning} (SSL) is a very large and active area of research, and we do not profess to cover all it; for a recent survey on SSL, we refer the reader to \citet{Engelen2019ASO}.

SSL estimates the distribution $P(\mathcal{X}=x)$ using a large amount of \textit{unlabeled}, to reduce the \textit{annotated} data requirements. It makes strong assumptions about the relation between $P(\mathcal{X}=x)$ and $P(\mathcal{Y}=y | \mathcal{X}=x)$ to reduce the number of labeled examples needed \citep{xiaojin2008semi}. Typically, these assumptions take the following forms:

\begin{itemize}
    \item \textbf{Smoothness:} Points that are close to each other are more likely to share a label. More formally, every two adjacent samples $x, x'$ should have similar labels. 
    \item \textbf{Cluster-ability:} Data tend to form discrete clusters where points belonging to the same cluster are more likely to share a label. Thus, the decision boundary can only pass through low-density areas in the feature space.
    \item \textbf{Manifold:} Data lies approximately on a manifold of a much lower dimension than the input space. Thus, when considering low-dimensional manifolds of the input space, any data points on the same manifold should have the same label. 
\end{itemize}

All three assumptions can be seen as different definitions of inter-points similarity: The smoothness defines it as proximity in the input space, the cluster-ability assumes high-density areas contain similar data points, and the manifold states that points which lie on the same low-dimensional manifold are similar.  

Another important distinction in SSL is between inductive and transductive methods. The former yield a classification model to predict the label of a new example, like supervised learning ($f:\mathcal{X} \xrightarrow{} \mathcal{Y}$). The latter do not yield such a model, but instead directly provide predictions.
Transductive approaches are usually graph-based, while the inductive approaches can be further divided into \emph{unsupervised preprocessing}, \emph{intrinsically semi-supervised} and \emph{wrapper methods} \citep{Engelen2019ASO}.

One popular way of using the \textit{unsupervised preprocessing} approach is to use the knowledge on $P(\mathcal{X}=x)$ to extract useful features in a lower dimension than the original dimension of $\mathcal{X}$ and thus reduce the learning complexity. This includes learning a representation using an auto-encoder model \citep{Vincent2008ExtractingAC} or applying a dimensionality reduction method like PCA \citep{alaiz2017convex}.

Under the inductive approach, it is also possible to use an \textit{intrinsically semi-supervised} model like semi-supervised SVM, which changes the optimization target to find a decision boundary with maximal-margin from both labeled and unlabeled points (e.g., using SVM) \citep{Vapnik1998StatisticalLT}. 
This can also be applied to neural networks by adding a form of regularization over the unlabeled data \citep{Rasmus2015SemisupervisedLW}.

In \textit{wrapper} methods, a model is initially trained from the available set of examples \citep{Triguero2013SelflabeledTF, lee2013pseudo}. It then makes predictions on the unlabeled dataset. The model’s pseudo-labels are added as labeled data for the next iteration of supervised learning. This process is repeated until convergence.


\begin{tcolorbox}[colback=blue!2!white,colframe=blue!50!black,title=Data programming]
Integrate multiple weak heuristics regarding the labeling process $f: \mathcal{X} \xrightarrow{} \mathcal{Y}$ to create noisy labels. 
\end{tcolorbox}

{\bf Data programming} is a paradigm for the programmatic creation of training sets. 
In data programming, users express weak supervision strategies or domain heuristics as labeling functions (LFs), which are programs that label subsets of the data \citep{ratner2016data}. Importantly, LFs are imprecise and can contradict each other, resulting in noisy labels. By explicitly representing the labeling process $f: \mathcal{X} \xrightarrow{} \mathcal{Y}$ as a generative model, data programming aims to ``denoise'' the generated training set.

For example, in spam-detection, potential LFs would return ``spam'' if the email contains a URL or a money transfer request, and ``no-spam'' if coming from someone in your contact list. These functions alone achieve poor performance; however, similar to ensemble methods (where a group of weak learners comes together to form a strong one with superior accuracy), the strength of data programming is in the combination of many weak heuristics. 

\begin{figure*}[h]
	\centering
  	\includegraphics[width=0.97\linewidth]{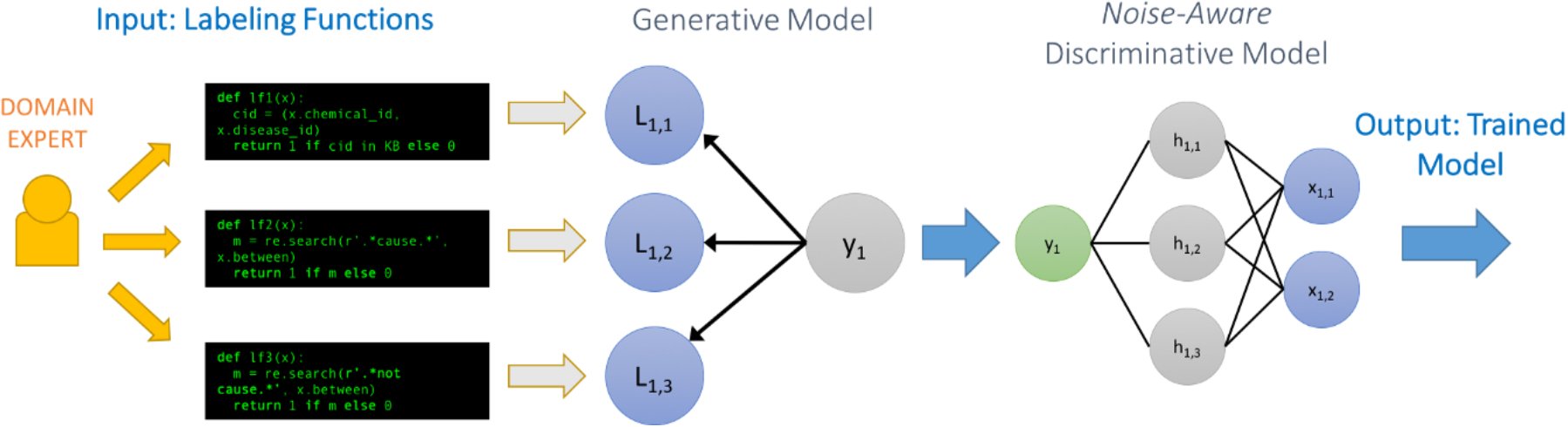}
  	\caption{Illustration of Snorkel's pipeline: first, the domain expert create noisy  labeling functions (LFs). A generative model learns to resolve and model the output of these LFs. The model's output is the input to a discriminative model. Image reproduced from \url{https://towardsdatascience.com/snorkel-a-weak-supervision-system-a8943c9b639f}
  \label{fig:Snorkel}}
\end{figure*}

A popular system for data programming is Snorkel \citep{ratner2019snorkel}. 
It applies the (noisy) LFs to the data and estimates their accuracy and correlations, using only their agreements and disagreements. This information is then used to reweight and combine LF predictions to output probabilistic noise-aware training labels.
This process is presented in Figure \ref{fig:Snorkel}.

\begin{tcolorbox}[colback=blue!2!white,colframe=blue!50!black,title=Expectation regularization]
Using prior knowledge regarding the proportion of the different labels in sub-groups of the data to create noisy labels.
\end{tcolorbox}

Prior knowledge regarding labels' \emph{proportion} in various sub-groups of the data, makes it possible to  automatically  create noisy labels in a process called \textbf{expectation regularization} (learn from label proportions) 
\citep{quadrianto2009estimating, mann2007simple, wang-manning-2014-cross}. 


This estimation process relies on uniform convergence properties of the expectation operator. It uses empirical means of the sub-groups to approximate expectations with respect to a group's distribution. The latter is then used to compute expectations with respect to a given label, and finally, the conditional means on the label distribution are used to estimate the conditional group means. 

A recent work in this area is \emph{ballpark learning}, which relaxes the assumption of known label proportions, assuming instead soft constraints on proportions within and between groups of instances (e.g., ``the percentage of spam in emails mentioning a certain word is between $k_{low}$ and $k_{high}$'', or ``emails containing a link have at least $k\%$ more spam than emails without links'') \citep{hope2016ballpark, hope2018ballpark}. Ballpark learning  learns a model that labels individual instances while satisfying these soft, noisy constraints.


\subsubsection{Noisy Supervision from External Datasets}
  \label{sec:relations}
  
It is sometimes possible to take advantage of preexisting datasets to get a noisy supervision signal.


\begin{tcolorbox}[colback=blue!2!white,colframe=blue!50!black,title=Distant supervision]
Use a preexisting database to collect examples for the desired relation. These examples are then used to automatically generate labeled training data.
\end{tcolorbox}

\textbf{Distant supervision} is a popular method to use existing datasets.
In distant supervision, a model is learned given a labeled training set, as in ``standard''
supervised ML, but the training data is weakly labeled (i.e., labeled automatically, based on heuristics or rules).  

For example, \citet{mintz2009distant} used Freebase, a large, unlabeled, semantic database, to provide distant supervision for relation extraction. The intuition is that any sentence that contains a pair of entities with a known Freebase relation is likely to express that relation in some way \citep{lin2015learning, fader2011identifying, hoffmann2011knowledge, dong2014knowledge, riedel2010modeling, zeng2014relation}. For example, each pair of ``Barack Obama'' and ``Michelle Obama'' that appear in the same sentence can be extracted as a positive example for the marriage relation. Due to the potentially large number of sentences that contain a given entity pair, it is possible to extract and combine noisy features for the labeling process. Based on these  semantic signals, \citet{mintz2009distant} were able to use 116 million unlabeled instances. 


\begin{tcolorbox}[colback=blue!2!white,colframe=blue!50!black,title=Incidental supervision]
Exploit weak signals that exist in data independently of the task at hand.
\end{tcolorbox}

The \textbf{incidental supervision} framework is based on the idea that informative cues for a task could exist in datasets that were not constructed with this task in mind. 
For example, suppose we want to infer gender from first names. One could use Wikipedia, which was not created for this task. The incidental signal would be pronouns and other gender indicators appearing in the first paragraph of Wikipedia pages about people with that first name.  This signal is correlated to the task at hand and (together with other signals and inferences), could be used for supervision, reducing the need for annotations.


Incidental supervision does not assume knowledge about the labeling process \citep{klementiev2007unsupervised, roth2017incidental}. Moreover, incidental signals can be noisy, partial, or only weakly correlated with the target task, and still be used to provide supervision and facilitate learning. \redd{Note that the notion of supervision here is different from
that of distant supervision: In distant supervision, the model learns in the standard supervised learning way, but the training set is labeled automatically, based on heuristics. In incidental supervision, a complete training set might never exist.}

Context-sensitive spelling and grammar checking is a task that has been relying on incidental supervision for over $20$ years now \citep{golding1999winnow}. Under the assumption that most edited textual resources (books, newspapers, Wikipedia) do not contain many spelling and grammar errors, these methods generate contextual representations for words, punctuation marks and phenomena such as agreements. These representations are then used to identify mistakes and correct them in a context-sensitive manner \citep{golding1999winnow, rozovskaya2014building}.

An unintentional example for the power of incidental signals comes from image processing, where the task of gender detection based on the iris texture was considered to be solved with great accuracy (over $80\%$ for most papers and an impressive score of $99.5\%$ reported by \citet{Alrashed2013FacialGR}) \citep{thomas2007learning, lagree2011predicting, bansal2012svm, tapia2014gender, da2015exploring, tapia2016gender}. However, it was later discovered that most models did not detect a person's gender; rather, they detected the use of cosmetic mascara, which is a much easier task and is indeed correlated with the original assignment \citep{kuehlkamp2017gender}. Thus, although unintentionally, this finding emphasized the potential of using incidental cues. 

%% file: sections/conclutions.tex
The dominant paradigm in ML today is creating large, task-specific datasets (often using crowdsourcing).  
In this review we devise a taxonomy for alternative ways to tackle the data bottleneck problem. 
The taxonomy aims to bring order to the the various methods suggested across different sub-fields, as well as making it easier to identify underlying assumptions and potential new directions. Identifying assumptions is essential for breaking them -- and breaking assumptions is an established technique for encouraging creativity and innovation. 

For example, surveying the taxonomy, several common assumptions that stand out are that samples tend to be representative of the data, that we have information about $X$ and $Y$ conjointly, and that each example has exactly one correct label. This raises the prospect of new learning settings (e.g., what if we only have knowledge about the distributions of data points $P(\mathcal{X}=x)$ and labels $P(\mathcal{Y}=y)$, separately?), and of new ways to aggregate multiple (correct but different) labels. 

\redd{We note that our taxonomy covers widely diverse techniques, making very different assumptions. Ultimately, we expect that choosing a technique will often boil down to what the practitioner has access to (that is, which assumptions are met). For example, in multi-task learning the practitioner not only possess labeled data for their task, but also for several related tasks; in data programming, they have no (or very few) labels for their task, but posses some partial knowledge about the labeling process; in curriculum learning, they know something about the hardness of data points; and so on.}

\redd{We further wish to point out that it is not always obvious whether a method's assumptions are met in practice, or to estimate which method is better suited for a specific use case. The answer might depend on many factors, such as the inherent difficulty of the concept one wishes to learn, biases in the data, or the manual effort needed to obtain high-quality input for the different methods. For example, in methods using weak labeling, the tradeoff between implementation speed and accuracy for different weak labels is often not clear in advance.}

In addition to the inherent difficulty of collecting large datasets, we note that there are growing concerns about such datasets, including environmental costs, financial costs, opportunity costs, and more \cite{bender2021dangers, schwartz2019allen}. We also note that large datasets are still prone to fitting artifacts \citep{mrqa_pres2019}, and that several recent methods have attempted to  address the recurring challenges of the annotation artifacts and human biases found in many existing datasets \citep{zellers2018swag, gururangan2018annotation}.

In conclusion, ML has made tremendous progress using large datasets, but they are not a panacea for all problems. Our hope is that this paper will encourage re-thinking about current annotation-heavy approaches.